\documentclass[10pt,twocolumn,letterpaper]{article}

\usepackage{cvpr}
\usepackage{times}
\usepackage{epsfig}
\usepackage{graphicx}
\usepackage{amsmath}
\usepackage{amssymb}

\usepackage{multirow}
\usepackage{algorithm}
\usepackage{algorithmic}

\graphicspath{{./figs/}}

\cvprfinalcopy 

\def\cvprPaperID{4043} 

\begin{document}

\title{Pyramidal Person Re-IDentification via Multi-Loss Dynamic Training}


\author{Feng Zheng\thanks{Major work was done when the author worked in YouTu Lab.}\\
Department of Computer Science and Engineering\\
Southern University of Science and Technology\\
Shenzhen 518055, China\\
\and
Cheng Deng\\
School of Electronic Enigineering\\
Xidian University\\
Xi'an 710071, China
\and
Xing Sun\thanks{Corresponding Author: winfredsun@tencent.com}, Xinyang Jiang, Xiaowei Guo, \\Zongqiao Yu, Feiyue Huang\\
YouTu Lab Tencent, Shanghai, China
\and
Rongrong Ji\thanks{Corresponding Author: rrji@xmu.edu.cn}\\
School of Information Science and Engineering,\\
Xiamen University, Xiamen, China\\
Peng Cheng Laboratory, Shenzhen, China
}

\maketitle
\thispagestyle{empty}


\begin{abstract}

Most existing Re-IDentification (Re-ID) methods are highly dependent on precise bounding boxes that enable images to be aligned with each other. However, due to the challenging practical scenarios, current detection models often produce inaccurate bounding boxes, which inevitably degenerate the performance of existing Re-ID algorithms. In this paper, we propose a novel coarse-to-fine pyramid model to relax the need of bounding boxes, which not only incorporates local and global information, but also integrates the gradual cues between them. The pyramid model is able to match at different scales and then search for the correct image of the same identity, even when the image pairs are not aligned. In addition, in order to learn discriminative identity representation, we explore a dynamic training scheme to seamlessly unify two losses and extract appropriate shared information between them. Experimental results clearly demonstrate that the proposed method achieves the state-of-the-art results on three datasets. Especially, our approach exceeds the current best method by $9.5\%$ on the most challenging CUHK03 dataset.

\end{abstract}

\section{Introduction}

Person Re-IDentification (Re-ID) aims to associate the images of the same person captured at different physical sites, facilitating cross-camera tracking techniques used in vision-based smart retail and security surveillance. In general, person Re-ID is considered to be the next high-level task after a pedestrian detection system, so the basic assumption of Re-ID is that the detection model can provide precise and highly-aligned bounding boxes. Despite the recent great progress, there are limited room for the performance improvement of existing methods due to the problems with part-based models and the difficulties in training.

\begin{figure}[t]
\vspace{-3mm}
\begin{center}
   \includegraphics[width=0.9\linewidth]{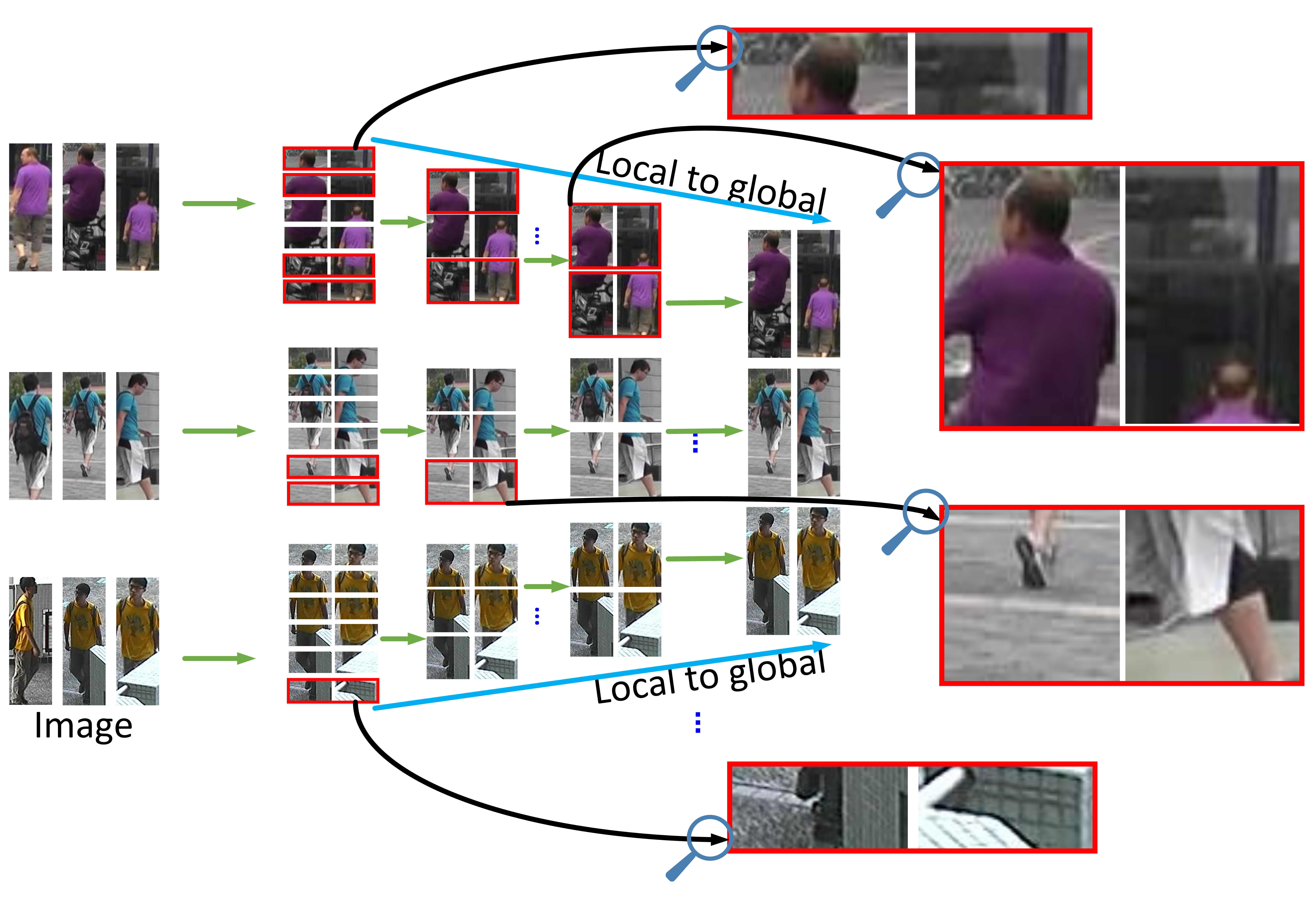}
\end{center}\vspace{-3mm}
   \caption{The examples of part-based matching at different scales, when bounding boxes are not aligned or parts of human body have been occluded. The red bounding box indicates that most cues in the two parts are varied. We can see that, in a finely partitioned way, a handful of horizontal stripes (left) cannot be well-matched due to different cues, while those stripes (right) in a more global view have more similar cues.}
\label{fig:motivation}
\vspace{-3mm}
\end{figure}

\textbf{Drawbacks of part-based models:} As it is well-known, part-based models can generally achieve promising performance in many computer vision tasks, because these models are potentially robust to some unavoidable challenges such as occlusion and partial variations. Actually, the performance of person Re-ID in the real-world applications is severely affected by these challenges. Thus, the recent proposed Part-based Convolutional Baseline (PCB) \cite{PCB_Sun2018} can achieve the state-of-the-art results. PCB is simple but very effective and even can outperform other learned part models. Nevertheless, in PCB, directly partitioning the feature map of backbone networks into a fixed number of parts strictly limits the capacities of further improving the performance. It has at least two major drawbacks, but not limited to: 1) The overall performance seriously depends on that a powerful and robust pedestrian detection model outputs a precise bounding box otherwise the parts cannot be well-aligned. However, in most cases of challenging scenes, current detection models are insufficient to do that. 2) The global information, which is also a very significant cue for recognition and identification, is completely ignored in this model whilst global features are normally robust to the subtle view changes and internal variations. Several examples are illustrated in Fig. \ref{fig:motivation} to show that the parts of diverse scales are equivalently important for matching.

\textbf{Difficulties of multi-loss training:} Recent studies \cite{Kumar2010,Kendall2018} demonstrate that multi-task learning has the capabilities to achieve advanced performance by extracting appropriate shared information between tasks. Without loss of generality, the terms ``loss'' and ``task'' will be used alternatively. In fact, many existing Re-ID methods \cite{MGN_Wang2018} also benefit from the multi-loss scheme to improve the performance. Generally, most multi-task methods choose to weight the losses using balancing parameters which are fixed during entire training process. 1) The performance highly relies on an appropriate parameter but choosing an appropriate parameter is undoubtedly a labor-intensive and tricky work. 2) The difficulties of different tasks actually change when the models are updated gradually, resulting in really varied appropriate parameters for different iterations. 3) More importantly, sampling strategies for different losses are generally diverse due to the specific considerations. For example, hard sample sampling for triplet loss would suppress the role of another task of identification loss.

To address above problems, in this paper, we specifically propose a novel coarse-to-fine pyramidal model based on the feature map extracted by a backbone network for person re-identification. First, the pyramid is actually a set of 3-dimensional sub-maps with a specific coarse-to-fine architecture, in which each member captures the discriminative information of different spatial scales. Then, a convolutional layer is used to reduce the dimension of features for each separated branch in the pyramid. Third, for each branch, the identification loss of a softmax function is independently applied to a fully connected layer which considers the features as the input. Furthermore, the features of all branches will be concatenated to form an identity representation, for which a triplet loss is defined to learn more discriminative features. To smoothly integrate the two losses, a dynamic training scheme with two sampling strategies is explored to optimize the parameters of deep neural networks. Finally, the learned identity representation will be used for person image matching, retrieval and re-identification.

In summary, the contribution of this paper is three-fold: 1) To relax the assumption of requiring a strong detection model, we propose a novel coarse-to-fine pyramid model that not only incorporates local and global information, but also integrates the gradual cues between them. 2) To maximally take advantage of different losses, we explore a dynamic training scheme to seamlessly unify two losses and extract appropriate shared information between them for learning discriminative identity representation. 3) The proposed method achieves the state-of-the-art results on the three datasets and most significantly, our approach exceeds the current best method by $9.5\%$ on dataset CUHK03.

\section{Related Work}

Most existing Re-ID methods either consider the local parts of person images or explore the global information. Some methods \cite{MultiLoss_Li2017,MGN_Wang2018} aware of that integrating the local and global features can improve the performance but the information between them is also ignored. We observe that those cues in the transition process are significant as well.

\textbf{Part-based algorithms:}
By performing bilinear pooling in a more local way, an embedding can be learned, in which each pooling is confined to a predefined region \cite{MultiRegion_Ustinova2015}. 
Inspired by attention models, in \cite{HydraPlus_Liu2017,HA-CNN_Li2018,DuATM_Si2018}, the attention-based deep neural networks are proposed to capture multiple attentions and select multi-scale attentive features.
Similarly, Zhao \emph{et al.} \cite{PAR_Zhao2017} explore a deep neural network method to jointly model body part extraction and representation computation, and learn model parameters.
Based on a L2 distance, \cite{MultiLoss_Li2017} formulates a method for joint learning of local and global feature selection losses particularly designed person Re-ID.
In \cite{PDC_Su2017}, a pose-driven deep convolutional model, which leverages the human part cues to alleviate the pose variations, is designed to learn feature extraction and matching models.
Furthermore, both the fine and coarse pose information of the person \cite{PSE+ECN_Sarfraz2018} are incorporated to learn a discriminative embedding.
A part loss is proposed in \cite{PartLoss_Yao2017}, which automatically detects human body parts and computes the person classification loss on each
part separately.
Chen \emph{et al.} \cite{MultiScale_Chen2017} develop a CNN-based appearance model to jointly learn scale-specific features and maximize multiscale feature fusion selections.
Several part regions are first detected and then deep neural networks are designed for representation learning on both the local and global regions \cite{GLAD_Wei2017}.
Part-based Convolutional Baseline (PCB) \cite{PCB_Sun2018} outputs a convolutional descriptor consisting of several part-level features and then a refined part pooling method is used to re-assign outliers in the parts. 
Based on PCB, Multiple Granularity Network (MGN) \cite{MGN_Wang2018} explores a branch for global features and two branches for local representations for person re-identificaiton.

\begin{figure*}[t]
\vspace{-3mm}
\begin{center}
   \includegraphics[width=0.9\linewidth]{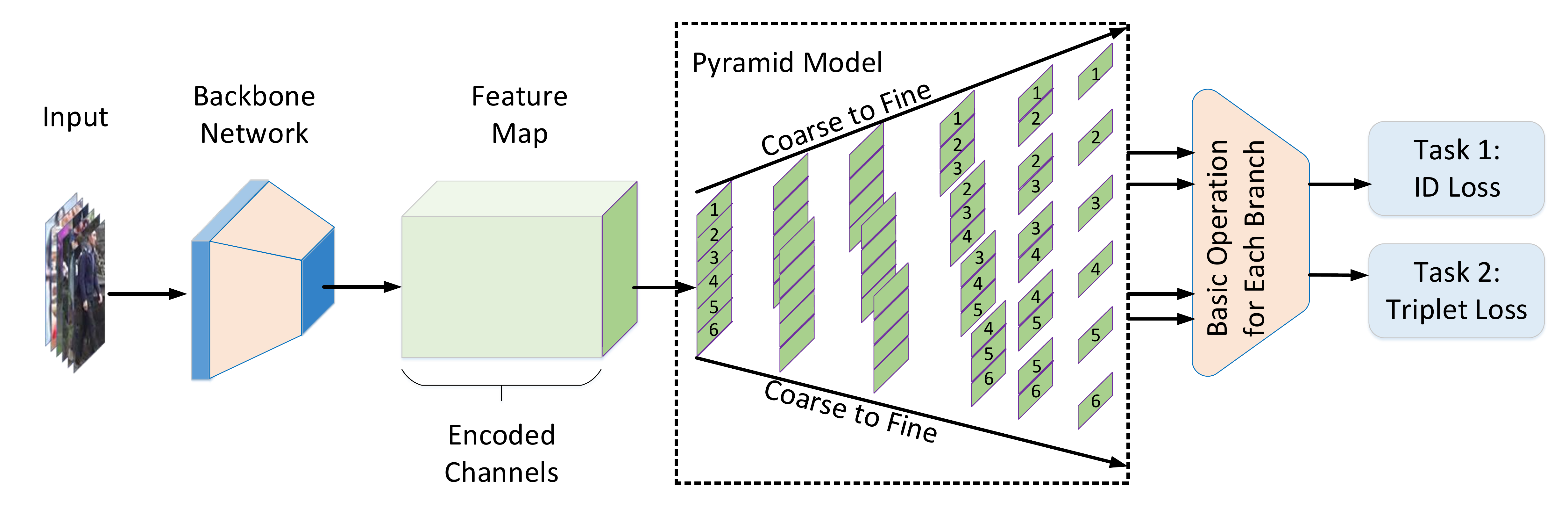}
\end{center}\vspace{-3mm}
   \caption{The architecture of our proposed pyramidal model for person re-identification. For better layout, only the spatial profile of the member branch in the pyramid is shown, which is originally 3-dimensional tensors. We assume that the original feature map is divided into 6 basic sub-maps, while other number of sub-maps can be used as well. The branch always consists of several consecutive basic sub-maps and the basic operation for each branch will be given in Fig. \ref{fig:pyramid_branch}.}
\label{fig:pyramid_model}
\vspace{-3mm}
\end{figure*}

\textbf{Non part-based methods:}
Recently, a completely synthetic dataset \cite{SOMAnet_Barbosa2017} and some adversarially occluded samples \cite{AOS_Huang2018} are constructed to train the re-identification model.
In \cite{SVDNet_Sun2017}, singular vector decomposition is used to iteratively integrate the orthogonality constraint in CNN training for image retrieval.
A pedestrian alignment network \cite{PAN_Zheng2018} is built to learn discriminative embedding and pedestrian alignment without extra annotations. 
Geng \emph{et al.} \cite{Transfer_Geng2016} propose a number of deep transfer learning models to address the data sparsity problem and transfer knowledge from auxiliary datasets.
\cite{Triplet_Hermans2017} shows that a plain CNN with a triplet loss can outperform most recent published methods by a large margin.
Learning binary representation for fast matching \cite{Zheng2016,Zheng2018} is also a promising direction of object re-identification.
In \cite{GSRW_Shen2018}, A group-shuffling random walk network is proposed to refine the probe-to-gallery affinities based on gallery-to-gallery affinities.
The ``local similarity'' metrics for image pairs are learned with considering dependencies from all the images in a group, forming ``group similarities'' in \cite{DNN-CRF_Chen2018}.

\textbf{Multi-task learning:}
Self-paced learning \cite{Kumar2010} and focal loss \cite{Lin2017} both train models by diversely weighting the samples in different learning stages. Inspired by this, in \cite{Li2017}, a task-oriented regularizer is designed to jointly prioritize both tasks and instances. 
The multiple loss functions \cite{Kendall2018} are weighted by considering the uncertainty of tasks in both classification and regression settings.
Moreover, a routing network consisting of two components is introduced to dynamically select different functions in response to the input \cite{Rosenbaum2018}.
While, Chen \emph{et al.} \cite{Chen2018} propose a gradient normalization algorithm that automatically balances training in deep multitask models by dynamically tuning gradient magnitudes. We learn the embedding for person Re-ID by simultaneously minimizing a list-wise metric loss and a classification loss with two types of sampling strategies.

\section{The Proposed Method}

\subsection{Coarse-to-Fine Pyramidal Model}

In this section, we propose a novel coarse-to-fine pyramidal model which can moderately relax the requirement of detection model and smoothly incorporate the global information, simultaneously. It is worth noting that, not only local and global information is integrated but the gradual transition process between them is also incorporated. 

\subsubsection{Pyramidal Branches}
Given a set of images $\mathbf{X} = \{I_1,\cdots,I_N\}$ containing persons captured by cameras in surveillance systems where $N$ is the number of images, the task of person Re-IDentification (Re-ID) is to associate the images of the same person at different times and locations. Our model is built on a feature map $M$ extracted by a backbone network $\mathcal{BN}$. Thus, we have a 3-dimensional tensor $M = \mathcal{BN}(I)$ of the size $C\times H \times W $, where $C$ is the number of encoded channels and $W$ and $H$ are the spatial width and height of the tensor, respectively.

First, we divide the feature map into $n$ number of parts according to the spatial height axis and thus each basic part has the size of $C\times (H/n) \times W $. Suppose that $H$ can be divisible by $n$. Thus, our pyramidal model is constructed according to the following rules: 1) In the bottom level ($l=1$) of the pyramid, there are $n$ number of branches in which one corresponds to a basic part. 2) The branches in higher level has one more adjacent basic part than that of previous lower level. 3) The sliding step for all levels is set to one. It means the number of branches in the current level is just one less than that of previous level. 4) In the top level ($l=n$) of the pyramid, there is only one branch which is just the original feature map $M$. Therefore, we assume that $\mathbf{P}\{l,k\}$ is the $k$th sub-map in the $l$th level of the pyramid model $\mathbf{P}$ defined as:
\begin{eqnarray}
\mathbf{P}=&&\{ M(1:C,st:ed,1:W):\\
&&st = (k-1)*H/n + 1,\nonumber\\
&&ed = (k-1)*H/n + l*H/n,\nonumber\\
&&l=1,\cdots,n, k=1,\cdots, n -l +1\},\nonumber
\end{eqnarray}
where $1:C$ means that all elements from index $1$ to index $C$ are selected. Obviously, $\mathbf{P}$ is a set of 3-dimensional sub-maps with a specific coarse-to-fine architecture, in which each member captures the discriminative information of different spatial scales. Moreover, the pyramidal model contains both the global feature map $M\in \mathbf{P}$ and the part-based model: PCB. While, it is easy to know that there are totally $\sum_{l=1}^{n} l$ number of components in $\mathbf{P}$ and the $l$th level has $n-l+1$ number of components, where the level index is in a fine-to-coarse fashion. The details of the proposed architecture are shown in Fig. \ref{fig:pyramid_model}.

\subsubsection{Basic Operations}

\begin{figure}[t]
\vspace{-3mm}
\begin{center}
   \includegraphics[width=0.9\linewidth]{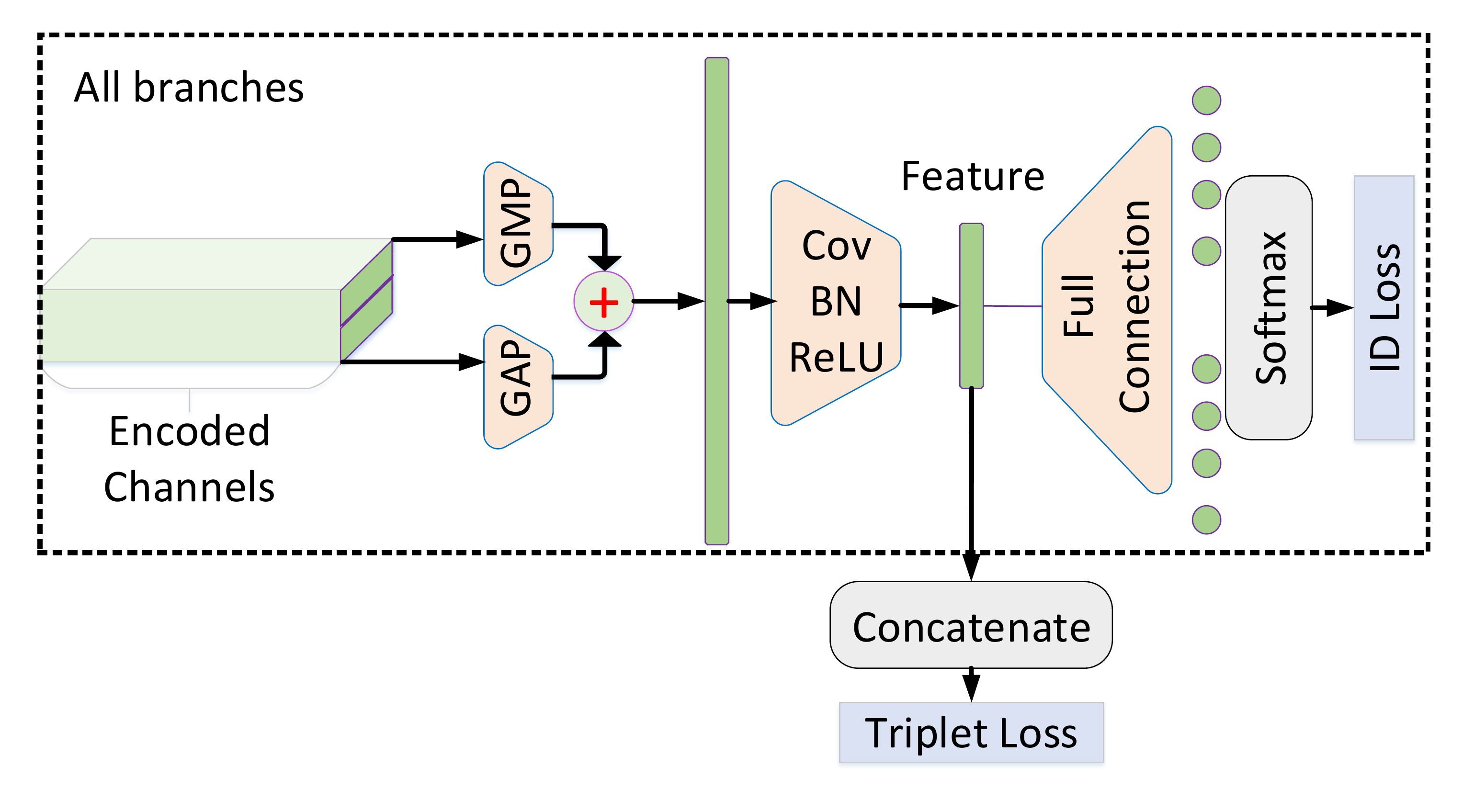}
\end{center}\vspace{-3mm}
   \caption{The illustration of basic operations for a branch consisting of two consecutive basic sub-maps, including a global maximum pooling, a global average pooling, a convolutional filter, a batch normalization, a ReLu activation and a linear fully connected layer. These operations for different branches will be executed independently and the features of all branches will be finally concatenated for the triplet loss.}
\label{fig:pyramid_branch}
\vspace{-3mm}
\end{figure}

For each branch $\mathbf{P}\{l,k\}$ in pyramid $\mathbf{P}$, first, a global maximum pooling (GMP) and a global average pooling (GAP) is separately executed to capture the statistical properties of different channels in the sub-maps. Then, the two statistical variables are added to form a vector with the same size of encoded channels. Third, a convolutional layer followed by a batch normalization and a ReLU activation is explored to reduce the dimension and produce a feature vector $x(l,k) \in R^D$ for the re-identification task. Simply, we denote $x(l,k) = \mathcal{BO}(\mathbf{P}\{l,k\})$. Fourth, to make the feature vector capable of sufficient discriminativity, a softmax based IDentification loss (ID loss) will be used for a fully connected layer which considers the feature map as the input. At the same time, a triplet loss will be imposed on a vector $x = (x(1,1)^T,\cdots,x(n,1)^T)^T$ which concatenates all the feature vectors of different branches in the pyramid $\mathbf{P}$. The basic operations for each branch will be executed independently with respect to different components in the pyramid. Finally, all the parameters will be learned by simultaneously minimizing the two losses in a dynamic way. A branch example consisting of two consecutive basic parts is illustrated in Fig. \ref{fig:pyramid_branch}. 

If assume that $f(\cdot)$ refers to all the operations in the embedding including $\mathcal{BN}$, $\mathbf{P}$ and $\mathcal{BO}$s, we can denote the feature as $x = f(I)$ simply. In the inference stage, the re-identification task will be achieved by ranking the distances $\{d(x,x_i):x = f(I),I_i\in \mathbf{G}\}$ between a query $I$ and a gallery $ \mathbf{G}$.

\subsection{Multi-Loss Dynamic Training}

Recent studies demonstrate that multi-task learning has the capabilities to achieve advanced performance by extracting appropriate shared information between tasks. The potential reason is that multiple tasks can benefit from each other by exploring the relatedness, leading to boosted generalization performance.

\subsubsection{Two Tasks}
To learn the discriminative features, we adopt two related tasks but emphasizing different aspects to learn the parameters of the embedding $f$, including an identification loss and a triplet loss. The first one is point-wise classification loss while the second one is for list-wise metric learning.

\textbf{Identification loss:} Generally, the identification loss is the same as the classification loss defined as:
\begin{eqnarray}\label{eq:id_loss}
L^{id} & = & \frac{1}{N_{id}} \sum_i \sum_{k,l} \mathcal{S}((W_c^{kl})^T x_i(l,k)) \\
& = & \frac{1}{N_{id}} \sum_i \sum_{k,l} - \log \frac{(W_c^{kl})^T x_i(l,k)}{\sum_j (W_j^{kl})^T x_i(l,k)},\nonumber
\end{eqnarray}
where $N_{id}$ is the number of used images, $c$ denotes the corresponding identity of the input image $I_i$, $\mathcal{S}$ is the softmax function and $W_c^{kl}$ is the weight matrix of the fully connected layer for $c$th identity in the $(l,k)$ branch. 

\textbf{Triplet loss:} Given a triplet of samples $(I,I_p,I_n)$ where $I$ and $I_p$ are of the same identity whilst $I$ and $I_n$ are the images for different identities, the aim of embedding is to learn a new feature space in which the distance between the sample pair $I$ and $I_p$ will be smaller than that between the pair $I$ and $I_p$. Intuitively, a triplet loss can be defined as:
\begin{eqnarray}\label{eq:triple_loss}
L^{tp} =  \frac{1}{N_{tp}} \sum_{I,I_p,I_n} [ d(f(I),f(I_p)) \\
- d(f(I),f(I_n))  + \delta]_+, \nonumber
\end{eqnarray}
where $\delta$ is a margin hyper-parameter to control the distance differences, $N_{tp}$ is the number of available triplets and $[\cdot]_+ = \max(\cdot,0)$ is the hinge loss.

\subsubsection{Dynamic Training}

The above two tasks are not novel and popularly used in various applications. While, how to integrate them is still an open problem. Actually, from the general perspective, most multi-task methods normally weight the tasks using balancing parameters and treat some tasks as the regularization items. In the learning stage, the balancing parameters are fixed during entire training process. 1) The performance strictly lies in an appropriate parameter but choosing an appropriate parameter is undoubtedly a labor-intensive and tricky work. 2) The difficulty of different tasks actually changes when the models are updated gradually, resulting in really varied appropriate parameters for different iterations. 

Furthermore, from the view of the re-identification task, to some extent, the two tasks are also conflicting when they are directly combined. On the one hand, effective triplets are rare, if the general random mini-batch sampling is used, making that the triplet loss contributes little in the learning procedure. This is because the number of identities is large but the number of images for each identity is small. On the other hand, to avoid the problem, we propose an ID-balanced sampling strategy to make sure triplets do exist in the mini-batches. However, this strategy suppresses the identification loss since fewer identities can be used in each mini-batch. Due to the sampling bias, it is possible that some images cannot be used all the time. Therefore, directly arithmetic weighting the losses would be very simple but obviously result in many difficulties in optimization. 

\textbf{Sampling:} To solve the problem, alternatively, we choose to dynamically minimize the two losses incorporating two sampling methods accordingly: random sampling and ID-balanced hard triplet sampling. Random sampling is easy to be implemented while ID-balanced hard triplet sampling is implemented according to the following steps. To build the effective triplets, we randomly select $8$ number of identities for each mini-batch, in which $8$ images of each identity are randomly chosen. Hence, this strategy definitely enables to use the hard positive/negative mining based on the largest intra-class (identity) distance and the smallest inter-class distance. However, the samples for different identities are unbalanced and those whose number of images is less than $8$ will never be used. 

\begin{figure}[t]
\vspace{-3mm}
\begin{center}
   \includegraphics[width=0.9\linewidth]{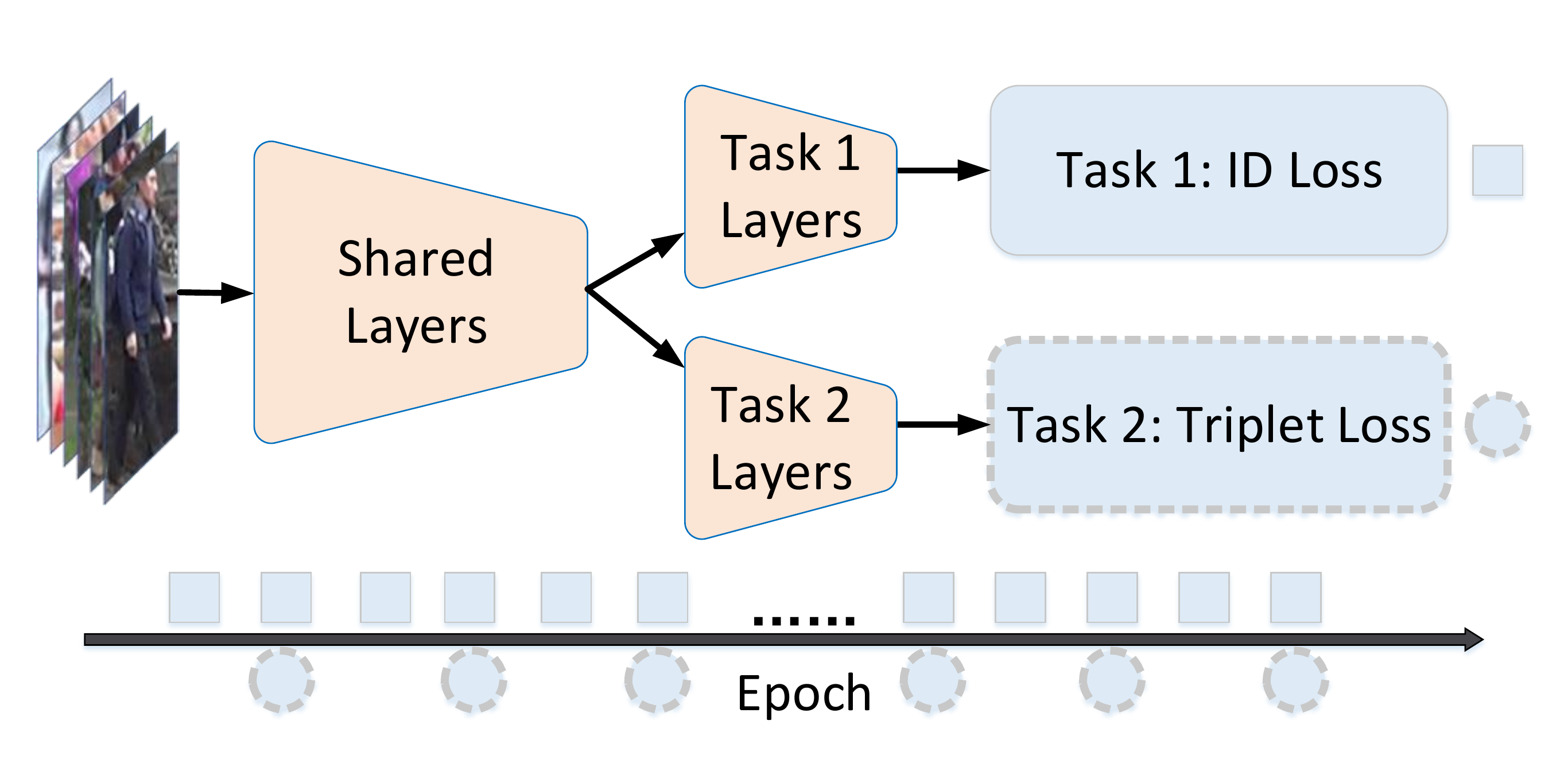}
\end{center}\vspace{-3mm}
   \caption{Dynamic training for two related tasks with two sampling strategies (\emph{i.e.}, ID-balanced hard triplet and randomized).}
\label{fig:multi-task}
\vspace{-3mm}
\end{figure}

\textbf{Dynamic weighting:} For each loss, we define a performance measure to estimate the likelihood of a loss reduction. Suppose $L_{\tau}^t$ be the average loss in the current training iteration $\tau$ for the task $t\in\{id,tp\}$. Thus, we can calculate $k_{\tau}^t$ to be an exponential moving average according to:
\begin{eqnarray}
k_{\tau}^t = \alpha L_{\tau}^t + (1 - \alpha)k_{\tau -1}^t,
\end{eqnarray}
where $\alpha\in [0,1]$ is a discount factor and $k_{-1}^t = L_{0}^t$. Based on the quantity $k_{\tau}^t$, we defined a probability to describe the likelihood of a loss reduction as:
\begin{eqnarray}
p_{\tau}^t  = \frac{\min\{k_{\tau}^t,k_{\tau-1}^t\}}{k_{\tau-1}^t}.
\end{eqnarray}
In case of loss increasing occasionally, the function $\min$ is used to normalize $p_{\tau}^t$ to be $1$. Obviously, $p_{\tau}^t = 1$ means the current optimization step didn't reduce the loss yet. The larger the value, the greater the probability that the optimization of the task $t$ steps into a local minimum. Similar to the Focal Loss which down-weights easier samples and concentrates on hard samples, we define a measure ($FL(\cdot)$) to weight the losses:
\begin{eqnarray}
FL(p_{\tau}^{t},\gamma) = -(1 - p_{\tau}^{t})^{\gamma}\log (p_{\tau}^{t}),
\end{eqnarray}
where $\gamma$ is used to control the focusing intensity. $FL(p_{\tau}^{t},\gamma)$ is designed to weight tasks and choose the desire loss to be optimized. Thus, the overall objective can be rewritten as:
\begin{eqnarray}\label{eq:overall_objective}
L = \sum_{t\in\{id, tp\}} FL(p_{\tau}^{t},\gamma) L_{\tau}^t.
\end{eqnarray}

Due to the different sampling strategies, we optimize the ID loss in Eq. \ref{eq:id_loss} with randomly selected mini-batches, when $FL(p_{\tau}^{id},\gamma)$ dominates the two tasks ($FL(p_{\tau}^{tp},\gamma)/FL(p_{\tau}^{id},\gamma) < \delta$). Thus, we start our dynamic optimization system from simply minimizing the ID loss. Actually, $FL(p_{\tau}^{id},\gamma)$ always dominates in the early optimization since each step can greatly reduce the ID loss. Moreover, because the model is currently in an immature status, all samples are equally difficult so that hard sampling-based triplet loss cannot play essential role for our optimization. This is similar to the scheme of self-paced (curriculum learning) leaning \cite{Kumar2010} in which easier samples are first trained and hard samples are considered later while here dynamically optimizing the two tasks plays the same role. In this case, the both losses $L_{\tau}^{t}: t\in\{id,tp\}$ in the objective Eq. \ref{eq:overall_objective} will be calculated.

When $FL(p_{\tau}^{tp},\gamma)$ dominates in the optimization, the overall objective \ref{eq:overall_objective} considering both Eq. \ref{eq:id_loss} and \ref{eq:triple_loss} will be directly optimized because ID-balanced sampling will not influence the use of ID loss. This optimization successfully avoids the tortuous parameter tuning and seamlessly incorporates the ideas of both ID-balanced hard triplet sampling and curriculum learning to further improve the performance. The flowchart of dynamic training is illustrated in Fig. \ref{fig:multi-task} and details of training are given in Algorithm \ref{alg:training}.

\begin{algorithm}[th]
\begin{center}
\caption{\hspace{5mm}Multi-Loss Dynamic Training}
\label{alg:training}
\begin{algorithmic}
 {\footnotesize
 \STATE \textbf{Input:}\hspace{0mm} Dataset $X$, pretrained backbone network $\mathcal{BN}$ and hyper-parameters $(n,\alpha,\delta,\gamma,D)$.
 \STATE \textbf{Output:}\hspace{0mm} The embedding function $x = f(I)$.
 \STATE \textbf{Initialization:}\hspace{0mm} 
 \STATE \hspace{0mm}Initiate the network parameters except for the backbone.
  \STATE \hspace{0mm}Set $p_{\tau}^{id} = 0$ and $p_{\tau}^{tp} = 1$.
\STATE \hspace{0mm}\textbf{for}\hspace{1mm} $\tau=1,\cdots,N_{\tau}$
\STATE \hspace{4mm}Calculate $FL(p_{\tau}^{tp},\gamma)$ and $FL(p_{\tau}^{id},\gamma)$.
\STATE \hspace{4mm}\textbf{if} $FL(p_{\tau}^{tp},\gamma)/FL(p_{\tau}^{id},\gamma) < \delta$
\STATE \hspace{8mm}Perform random mini-batch sampling.
\STATE \hspace{8mm}Forward though $\mathcal{BN}$ and construct the pyramid $\mathbf{P}$.
\STATE \hspace{8mm}Apply batch-normalization and calculate $x = f_{\tau}(I)$.
\STATE \hspace{8mm}Optimize the objective in Eq. \ref{eq:id_loss}.
\STATE \hspace{4mm}\textbf{else}
\STATE \hspace{8mm}Perform ID-balanced hard triplet sampling.
\STATE \hspace{8mm}Forward though $\mathcal{BN}$ and construct the pyramid $\mathbf{P}$.
\STATE \hspace{8mm}Apply batch-normalization and calculate $x= f_{\tau}(I)$.
\STATE \hspace{8mm}Optimize the objective in Eq. \ref{eq:overall_objective}.
\STATE \hspace{4mm}\textbf{end if}
\STATE \hspace{4mm}Backpropagate the gradients and update the parameters.
\STATE \hspace{0mm}\textbf{end for}
}
\end{algorithmic} 
\end{center}
\end{algorithm}

\section{Experiment}

To verify the proposed method, we test it on three popular used person re-identification datasets: Market-1501 \cite{Zheng2015}, DukeMTMC-reID \cite{Ristani2016} and CUHK03 \cite{Li2014}.

\subsection{Experimental Setting}

\textbf{Implementation details:}
All images are resized into a resolution of $384\times 128$ which is the same as that of PCB.
The ResNet model with the pretrained parameters on ImageNet is considered as the backbone network in our system.
For the feature map, the number of encoded channels is $2048$ while the feature will be reduced to a $128$-dimensional vector using a convolutoinal layer.
We set the number of basic parts to $6$ so there are $21$ branches in the pyramid according to construction rules.
The margin in the triplet loss is $1.4$ in all our experiments.
We select a mini-batch of $64$ images for each iteration.
Stochastic gradient descent (SGD) with two sampling strategies is used in our optimization, where the momentum and the weight decay factor is set to $0.9$ and $0.0005$, respectively. Totally, the proposed model will be trained $120$ epochs.
As for the learning rate, the initial learning rate is set to $0.01$ while the learning rate will be dropped by half every 10 epochs from epoch $60$ to epoch $90$. While, for the dynamic training, we set the parameters $\delta = 0.16$, $\alpha = 0.25$ and $\gamma = 2$, according to the suggestions in \cite{Lin2017}. All the experiments in this paper will follow the same setting.

\textbf{Evaluation metrics:}
To compare the re-identification performance of the proposed method with the existing advanced methods, we adopt the Cumulative Matching Characteristics (CMC) at rank-1, rank-5 and rank-10, and mean Average Precision (\textit{mAP}) on all the datasets. It is worth noting that all our results are obtained in a single-query setting and, more importantly, re-ranking algorithm is not used to improve the \textit{mAP} in all experiments.

\subsection{Datasets} 

\textbf{Market-1501:}
In this dataset, $32,668$ images of $1,501$ identities with annotated bounding boxes detected using the pedestrian detector of Deformable Part Model (DPM) are collected. View overlapping exists among different cameras, including $5$ high-resolution cameras, and a low-resolution camera. Following the setting of PCB, we divide the dataset into a training set with $12,936$ images of $751$ persons and a testing set of $750$ persons containing $3,368$ query images and $19,732$ gallery images.

\textbf{DukeMTMC-reID:}
Following the protocol \cite{Duke-reid-Zheng2017} of the Market-1501 dataset, this dataset is a subset of the DukeMTMC dataset specifically collected for person re-identification. In this dataset, $1,404$ identities appears in more than two cameras while $408$ (distractor) identities appears in only one camera. We divide the dataset into a training set of $16,522$ images with $702$ identities and a testing set which consists of $2,228$ query images of the other $702$ identities and $17,661$ gallery images of $702$ identities plus $408$ distractor identities.

\textbf{CUHK03:}
We follow the new protocol \cite{Zhong2017} similar to that of Market-1501, which splits the CUHK03 dataset into training set of $767$ identities and testing set of $700$ identities. From each camera, one image is selected as the query for each identity and the rest of images are used to construct the gallery set. This dataset has two ways of annotating bounding box including labelled by human or detected by a detector. The labelled dataset includes $7,368$ training, $1,400$ query and $5,328$ gallery images while detected dataset consists of $7,365$ training, $1,400$ query and $5,332$ gallery images.

\subsection{Comparison with State-of-the-Art Methods}

\begin{table}
\begin{center}
{\footnotesize
\begin{tabular}{l|ccc|c}
\hline
Method	&	\textit{mAP} 	&	\textit{rank 1}	&	\textit{rank 5}	&	\textit{rank 10}	\\
\hline\hline									
Pyramid-ours	&	\textbf{88.2}	&	\textbf{95.7}	&	\textbf{98.4}	&	\textbf{99.0} 	\\
\hline									
MGN \cite{MGN_Wang2018}	&	86.9 	&	95.7 	&	-	&	-	\\
PCB+RPP \cite{PCB_Sun2018}	&	81.6 	&	93.8 	&	97.5 	&	98.5 	\\
PCB \cite{PCB_Sun2018}	&	77.4 	&	92.3 	&	97.2 	&	98.2 	\\
GLAD* \cite{GLAD_Wei2017}	&	73.9 	&	89.9 	&	-	&	-	\\
MultiScale \cite{MultiScale_Chen2017}	&	73.1 	&	88.9 	&	-	&	-	\\
PartLoss \cite{PartLoss_Yao2017}	&	69.3 	&	88.2 	&	-	&	-	\\
PDC* \cite{PDC_Su2017}	&	63.4 	&	84.4 	&	92.7 	&	94.9 	\\
MultiLoss \cite{MultiLoss_Li2017}	&	64.4 	&	83.9 	&	-	&	-	\\
PAR \cite{PAR_Zhao2017}	&	63.4 	&	81.0 	&	92.0 	&	94.7 	\\
HydraPlus \cite{HydraPlus_Liu2017}	&	-	&	76.9 	&	91.3 	&	94.5 	\\
MultiRegion \cite{MultiRegion_Ustinova2015}	&	41.2 	&	66.4 	&	85.0 	&	90.2 	\\
\hline									
DML \cite{DML_Zhang2018}	&	68.8 	&	87.7 	&	-	&	-	\\
Triplet Loss \cite{Triplet_Hermans2017}	&	69.1 	&	84.9 	&	94.2 	&	-	\\
Transfer \cite{Transfer_Geng2016}	&	65.5 	&	83.7 	&	-	&	-	\\
PAN \cite{PAN_Zheng2018}	&	63.4 	&	82.8 	&	-	&	-	\\
SVDNet \cite{SVDNet_Sun2017}	&	62.1 	&	82.3 	&	92.3 	&	95.2 	\\
SOMAnet \cite{SOMAnet_Barbosa2017}	&	47.9 	&	73.9 	&	-	&	-	\\
\hline
\end{tabular}}
\end{center}
\caption{Comparison results ($\%$) on Market-1501 dataset at $4$ evaluation metrics: \textit{mAP}, \textit{rank 1}, \textit{rank 5} and \textit{rank 10} where the bold font denotes the best method. ``*'' denotes that the method needs auxiliary part labels. We divide other compared methods into two groups: the methods exploring part-based features and the methods extracting global features. Our proposed pyramid model achieves the best results on all the $4$ evaluation metrics.}\label{tbl:Market}
\end{table}
%

\begin{table}
\begin{center}
{\footnotesize
\begin{tabular}{l|cc}
\hline
Method	&	\textit{mAP} 	&	\textit{rank 1}	\\
\hline\hline					
Pyramid-ours	&	\textbf{79.0}	&	\textbf{89.0}	\\
\hline					
MGN  \cite{MGN_Wang2018}	&	78.4 	&	88.7 	\\
SVDNet \cite{SVDNet_Sun2017}	&	56.8 	&	76.7 	\\
AOS \cite{AOS_Huang2018}	&	62.1 	&	79.2 	\\
HA-CNN \cite{HA-CNN_Li2018}	&	63.8 	&	80.5 	\\
GSRW \cite{GSRW_Shen2018}	&	66.4 	&	80.7 	\\
DuATM \cite{DuATM_Si2018}	&	64.6 	&	81.8 	\\
PCB+RPP \cite{PCB_Sun2018}	&	69.2 	&	83.3 	\\
PSE+ECN \cite{PSE+ECN_Sarfraz2018}	&	75.7 	&	84.5 	\\
DNN-CRF \cite{DNN-CRF_Chen2018}	&  69.5 & 84.9 \\
GP-reid \cite{GP-reid_Xiong2018}	&	72.8 	&	85.2 	\\
\hline
\end{tabular}}
\end{center}
\caption{Comparison results ($\%$) on DukeMTMC-reID dataset. The results of our proposed pyramid model at \textit{rank 5} and \textit{rank 10} are $94.7\%$ and $96.3\%$, respectively, which achieve the state-of-the-art.}\label{tbl:DukeMTMC}
\end{table}
%

\begin{table}
\begin{center}
{\footnotesize
\begin{tabular}{l|cc|cc}
\hline
\multirow{2}{*}{Method}	&	\multicolumn{2}{c|}{Labelled}	&			\multicolumn{2}{c}{Detected}			\\
\cline{2-5}									
	&	\textit{mAP} 	&	\textit{rank 1}	&	\textit{mAP} 	&	\textit{rank 1}	\\
\hline\hline									
Pyramid-ours	&	\textbf{76.9}	&	\textbf{78.9}	&	\textbf{74.8}	&	\textbf{78.9}	\\
\hline									
MGN {\cite{MGN_Wang2018}}	&	67.4 	&	68.0 	&	66.0 	&	68.0 	\\
PCB+RPP {\cite{PCB_Sun2018}}	&	-	&	-	&	57.5	&	63.7	\\
MLFN {\cite{MLFN_Chang2018}}	&	49.2 	&	54.7	&	47.8	&	52.8	\\
HA-CNN {\cite{HA-CNN_Li2018}}	&	41.0 	&	44.4	&	38.6	&	41.7	\\
SVDNet {\cite{SVDNet_Sun2017}}	&	37.8 	&	40.9	&	37.3	&	41.5	\\
PAN {\cite{PAN_Zheng2018}}	&	35.0 	&	36.9	&	34	&	36.3	\\
IDE {\cite{IDE_Zheng2016}}	&	21.0 	&	22.2	&	19.7	&	21.3	\\
\hline
\end{tabular}}
\end{center}
\caption{Comparison results ($\%$) on CUHK03 dataset using the new protocol in \cite{Zhong2017}. For the labelled set, the results of our model at \textit{rank 5} and \textit{rank 10} are $91.0\%$ and $94.4\%$, respectively, while they are $90.7\%$ and $94.5\%$ for the detected dataset. This dataset is the most difficult one by comparing the average performance of all methods. The proposed pyramid model outperforms all the other state-of-the-art methods with large margins.}\label{tbl:CUHK03}
\end{table}
%

In this section, we compare the proposed method called ``Pyramid-ours'' with $26$ state-of-the-art methods, most of which is proposed in the last year, on the three datasets including Market-1501, DukeMTMC-reID and CUHK03. For the comparison of each dataset, we detail the following.

\textbf{Market-1501:}
For this dataset, we divide the compared methods into two groups: including part-based and global methods, and the comparisons are given in Tab. \ref{tbl:Market}. The results clearly show that local-based methods generally get better evaluation scores than that of these methods extracting global features only. The PCB is a convolutional baseline that motivates our approach, but we have improved performance by $10.8\%$ and $3.4\%$ on metrics \textit{mAP} and \textit{rank 1}, respectively. MGN considers multiple branches as well but ignores the gradual cues between global and local information. Our method achieves the same result with MGN on metric \textit{rank 1} but exceeds it $1.3\%$ on metric \textit{mAP}. In comparison, the performances of other algorithms are similar with each other on metric \textit{rank 10} but all of them are much worse on metrics \textit{mAP} and \textit{rank 1} than ours. 

\textbf{DukeMTMC-reID:} From Tab. \ref{tbl:DukeMTMC}, we can see that our method also achieves the best results on this dataset at both metrics \textit{mAP} and \textit{rank 1}. Among the compared methods, MGN is the closest method to our method score, but still below $0.6\%$ \textit{mAP} score. PSE+ECN which is a method using a pose-sensitive embedding and the re-ranking procedure also performs worse than ours ($75.7\%$ vs. $79.0\%$). Similar to the comparison on Market-1501 dataset, our pyramid model exceeds PCB+RPP $9.8\%$ and $5.7\%$ at metrics \textit{mAP} and \textit{rank 1}, respectively. We provide the achievements of our method at metrics \textit{rank 5} ($94.7\%$ ) and \textit{rank 10} ($96.3\%$ ) for comparison in the future.

\textbf{CUHK03:} 
This dataset is the most challenging dataset under the new protocol and the bounding boxes are annotated using two ways. While, from Tab. \ref{tbl:CUHK03}, we can see that our proposed approach has achieved the most outstanding results for these two annotation ways. On this datasst, the pyramid model outperforms all other methods at least $8.8\%$ and $10.9\%$, respectively.

\begin{figure}[t]
\vspace{-3mm}
\begin{center}
   \includegraphics[width=0.9\linewidth]{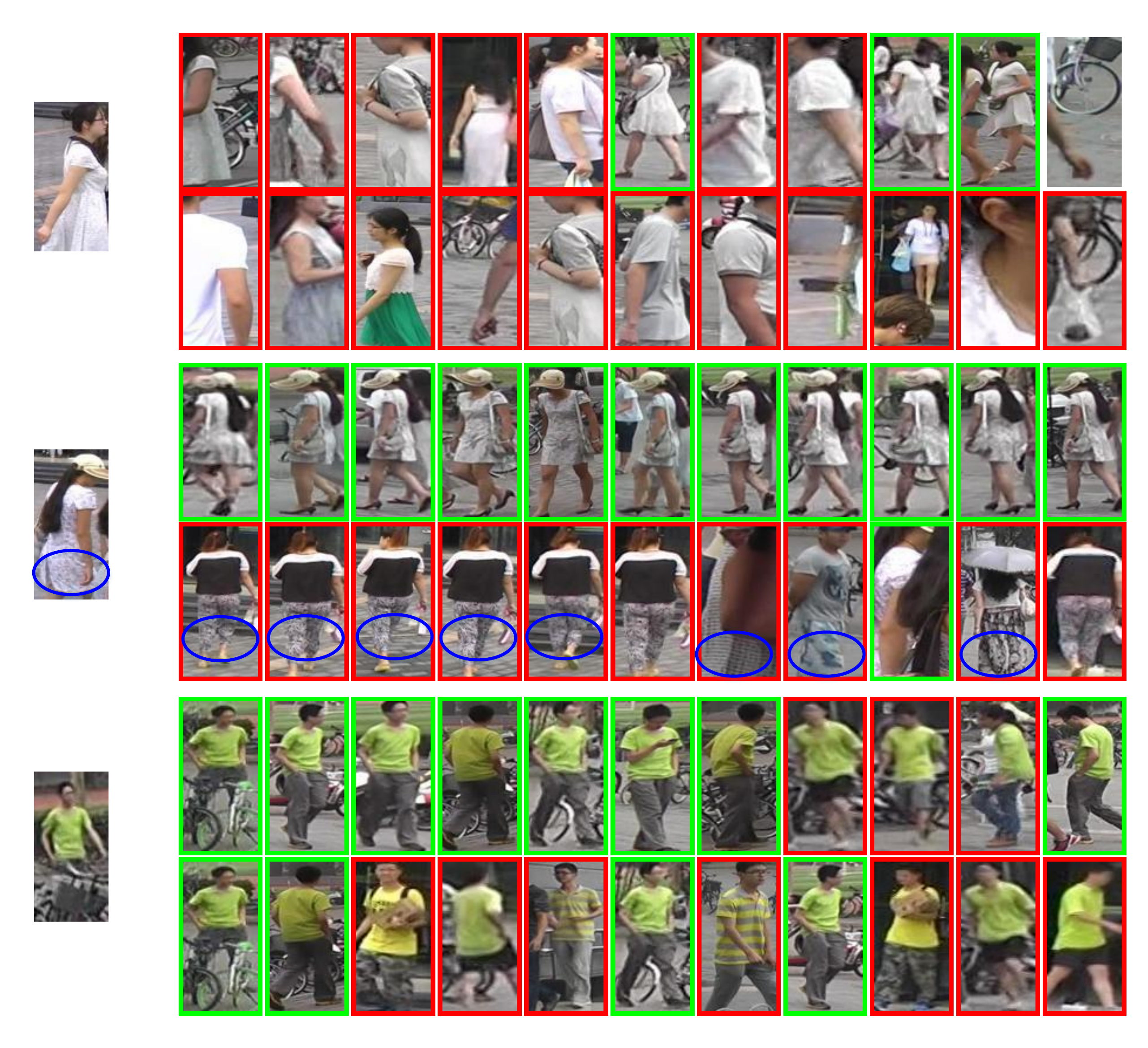}\\
Query~~~~~~~~~~~~~~~~~~~~Top $11$ retrieved images~~~~~~~~~~~~~~~~~~~
\end{center}
   \caption{Examples of retrieved images by two methods: Pyramid-ours and PCB, in case of the imprecise detection. For each query, the images in first row are returned by our proposed method while the images in second row are searched by PCB. The green/red rectangles indicate that images have the same/different identities as the query and the blue eclipse marks the similar contexts.}
\label{fig:comparison_pcb_pyramid}
\vspace{-3mm}
\end{figure}

Furthermore, on Market-1501 dataset, we compare our model with PCB using the same sampling strategy and some retrieved examples are shown in Fig. \ref{fig:comparison_pcb_pyramid}. We can see that the PCB cannot respond well to the challenge of inaccurate bounding boxes. Taking the first query as an example, our model is able to find three images of the same identity in the top $11$ results whilst PCB could not search anyone. While, from the second query, we can see that the lower-body parts (blue eclipse) of retrieved images match to the upper-body part of query, due to the imprecise detection.

In summary, our proposed pyramid using the novel multi-loss dynamic training can always be superior to all other existing advanced methods, no matter which evaluation metric is used. Through the comparative experiment on the three datasets, it is easy to know that CUHK03 dataset with the new protocol would be the most challenging one because all methods make worse results on it. However, our method can consistently outperform all other algorithms by a large margin. Therefore, we can conclude that our method particularly specializes in challenging problems.

\subsection{Component Analysis}

To further investigate every component in the pyramid model, we conduct comprehensive ablation studies on the performance of different sub-models. The results at the metrics: \textit{mAP}, \textit{rank 1}, \textit{rank 5} and \textit{rank 10} are shown in Tab. \ref{tbl:ablation} and each result is obtained with only one setting changed and the rest being the same as the default.

First, we merely use the part of branches in the pyramid to test the function. In the term ``Pyramid-000001'', the left number denotes whether the branches in low level are used or not while the rightest number is for the global branch. For example, ``Pyramid-000001'' means only the global branch in the highest level of the pyramid is used. From this table, we can obtain: 1) The local branches in the lower levels play more important roles than that of the global branches ($84.9\%$ vs. $82.1\%$). 2) The more branches we use, the better the performance. 3) The only global branch plus the proposed dynamic training strategy can achieve better results than that of PCA+RPP. It clearly shows the dynamic training strategy is able to improve the capacities of the model.

Second, the features of different dimensions are also analyzed. Compared to the default dimension $128$, the features of dimension $64$ and $256$ both achieve worse results. It shows that the redundant information plays negative influence to the performance while too short feature cannot provide sufficient discriminative cues. In summary, the performance is relatively $(\leq 1.3\%)$ stable with the change of the feature dimension. While, in the case of resource-limited application, $64$-dimensional feature is a more acceptable choice.

Finally, we fix the dynamic balance parameter to $0$ and alternately execute the two sampling strategies to train the identification loss. It means that the triplet loss will never be used in this experiment. In one step, mini-batch is selected using random sampling while ID-balance hard sampling is adopted in the next step. We can see that the overall performance is a little bit lower than that of the default setting of our proposed model but still much higher than that of PCB-RPP. It demonstrates the new pyramid model and the dynamic sampling strategy contribute most for the performance improvement.

\begin{table}
\begin{center}
{\footnotesize
\begin{tabular}{l|cccc}
\hline
Model	&	\textit{mAP} 	&	\textit{Rank 1}	&	\textit{Rank 5}	&	\textit{Rank 10}	\\
\hline\hline									
Pyramid-000001	&	82.1 	&	92.8 	&	97.3 	&	98.2 	\\
Pyramid-100000	&	84.9 	&	93.9 	&	97.6 	&	98.5 	\\
Pyramid-001111	&	86.7 	&	94.8 	&	98.4 	&	98.8 	\\
Pyramid-110011	&	87.2 	&	95.0 	&	98.1 	&	98.8 	\\
Pyramid-111100	&	87.5 	&	94.8 	&	98.3 	&	98.9 	\\
Feature-64	&	86.9 	&	94.5 	&	97.8 	&	98.6 	\\
Feature-256	&	87.8 	&	95.3 	&	98.2 	&	98.9 	\\
No triplet loss	&	86.5 	&	93.8 	&	97.5 	&	98.4 	\\
\hline									
Pyramid-ours	&	\textbf{88.2}	&	\textbf{95.7}	&	\textbf{98.4}	&	\textbf{99.0}	\\
PCB+RPP {\cite{PCB_Sun2018}}	&	81.6 	&	93.8 	&	97.5 	&	98.5 	\\
\hline
\end{tabular}}
\end{center}
\caption{Results ($\%$) of sub-models on Market-1501 dataset. In the term ``Pyramid-000001'', `0' means the corresponding level of pyramid is not used while `1' means that is used. ``Feature-64'' denotes the dimension of features for each branch is set $64$. ``No triplet loss'' refers to that only identification loss is optimized.}\label{tbl:ablation}
\end{table}
%

\section{Conclusion}

In this paper, we construct a coarse-to-fine pyramid model for person re-identification via a novel dynamic training scheme. Our model relaxes the requirement of detection models and thus achieves advanced results on benchmark. Specially, our model outperforms the existing best method by a large margin on CUHK03 dataset which is the most challenging dataset under the new protocol of \cite{Zhong2017}. It is worth noting that all our results are achieved in a single-query setting without using any re-ranking algorithms. In the future, it will be interesting to jointly learn the detection and re-identification models in an integrated framework. The two tasks are highly related and Re-ID models can be improved by means of attention maps in the detection models. Moreover, the middle layer features in the backbone can be incorporated into the proposed pyramid model as well to further improve the Re-ID performance.

\section*{Acknowledgement}
Our work was supported in part by the Science and Technology Innovation Committee Foundation of Shenzhen (Grant No. ZDSYS201703031748284), the Program for University Key Laboratory of Guangdong Province (Grant No. 2017KSYS008), the National Natural Science Foundation of China (Grant No. 61572388 and 61703327) and the Key R\&D Program-The Key Industry Innovation Chain of Shaanxi (Grant No. 2017ZDCXL-GY-05-04-02).

{\small
\bibliographystyle{ieee_fullname}
\bibliography{pyramid}
}

\end{document}